\documentclass[a4paper,french]{rnti}
\usepackage[T1]{fontenc}
\usepackage[frenchb]{babel}
\usepackage{graphicx}
\usepackage{caption}
\usepackage{subcaption}
\usepackage{multirow}
\usepackage{amsmath,amssymb,amsfonts}
\usepackage{multicol}
\usepackage{tabularx}
\usepackage{url}

\usepackage{listings}
\usepackage{latexsym}

\usepackage{fancyhdr}

\newcolumntype{C}{>{\centering\arraybackslash}X}

\makeatletter
\def\input@path{{./tex/}}
\graphicspath{{./graphics/}}

\titrecourt{USTEP: Structuration des logs}

\nomcourt{Vervaet et al.}

\titre{USTEP: Structuration des logs en flux grâce à un arbre de recherche évolutif}

\auteur{Arthur Vervaet\affil{1},
        Raja Chiky\affil{2},
        Mar Callau-Zori\affil{1}}

\affiliation{
    \affil{1}3DS Outscale, Saint-Cloud, France\\
    \affil{2}ISEP - Institut Supérieur d’Electronique de Paris, Issy-les-Moulineaux, France\\
    arthur.vervaet@outscale.com, raja.chiky@isep.fr\\ 
 }

\resume{%
Les registres de \textit{logs} enregistrent en temps réel des informations 
relatives à l'exécution d'un système informatique. Ceux-ci sont régulièrement consultés à 
des fins de développement ou de surveillance. 
Afin d'être utilisables plus facilement pour des tâches d'exploration automatique,
les messages des \textit{logs} doivent être structurés.
Cette étape se déroulant en amont des opérations d'analyses, elle
peut devenir un goulot d'étranglement temporel et influencer l'efficacité des méthodes en aval.
Dans ce papier, nous présentons USTEP, une méthode de 
structuration des \textit{logs} en flux. Basée sur un arbre de recherche évolutif,
USTEP est capable de découvrir et d'encoder efficacement de nouvelles règles de structuration.
Nous présentons ici une évaluation des performances de USTEP sur un panel de 13 jeux de données issus de 
systèmes différents. Nos résultats mettent en valeur la supériorité de notre approche en matière d'efficacité et 
de robustesse par rapport à d'autres méthodes de structuration en ligne.
}

\summary{%
Logs record valuable system information at runtime. 
They are widely used by data-driven approaches for development and 
monitoring purposes. Parsing log messages to structure their format is a 
classic preliminary step for log-mining tasks. As they appear upstream, 
parsing operations can become a processing time bottleneck for downstream 
applications. The quality of parsing also has a direct influence on their 
efficiency. 
Here, we propose USTEP, an online log parsing method based on an evolving tree 
structure. Evaluation results on a wide panel of datasets coming from 
different real-world systems demonstrate USTEP superiority in terms of 
both effectiveness and robustness when compared to other online methods. 
}

\begin{document}
%\layout

% DEBUT DE L'ARTICLE
\section{Introduction}\label{sec:intro}

\begin{figure}[h]
    \centering
    \begin{minipage}{.5\textwidth}
      \centering
      \includegraphics[width=0.9\linewidth]{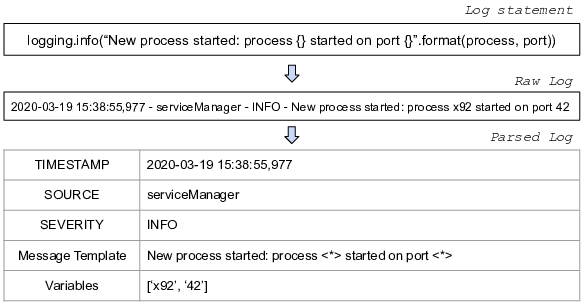}
      \captionof{figure}{Exemple de structuration de log}
      \label{fig:logparsing}
    \end{minipage}%
    \begin{minipage}{.5\textwidth}
      \centering
      \includegraphics[width=0.9\linewidth]{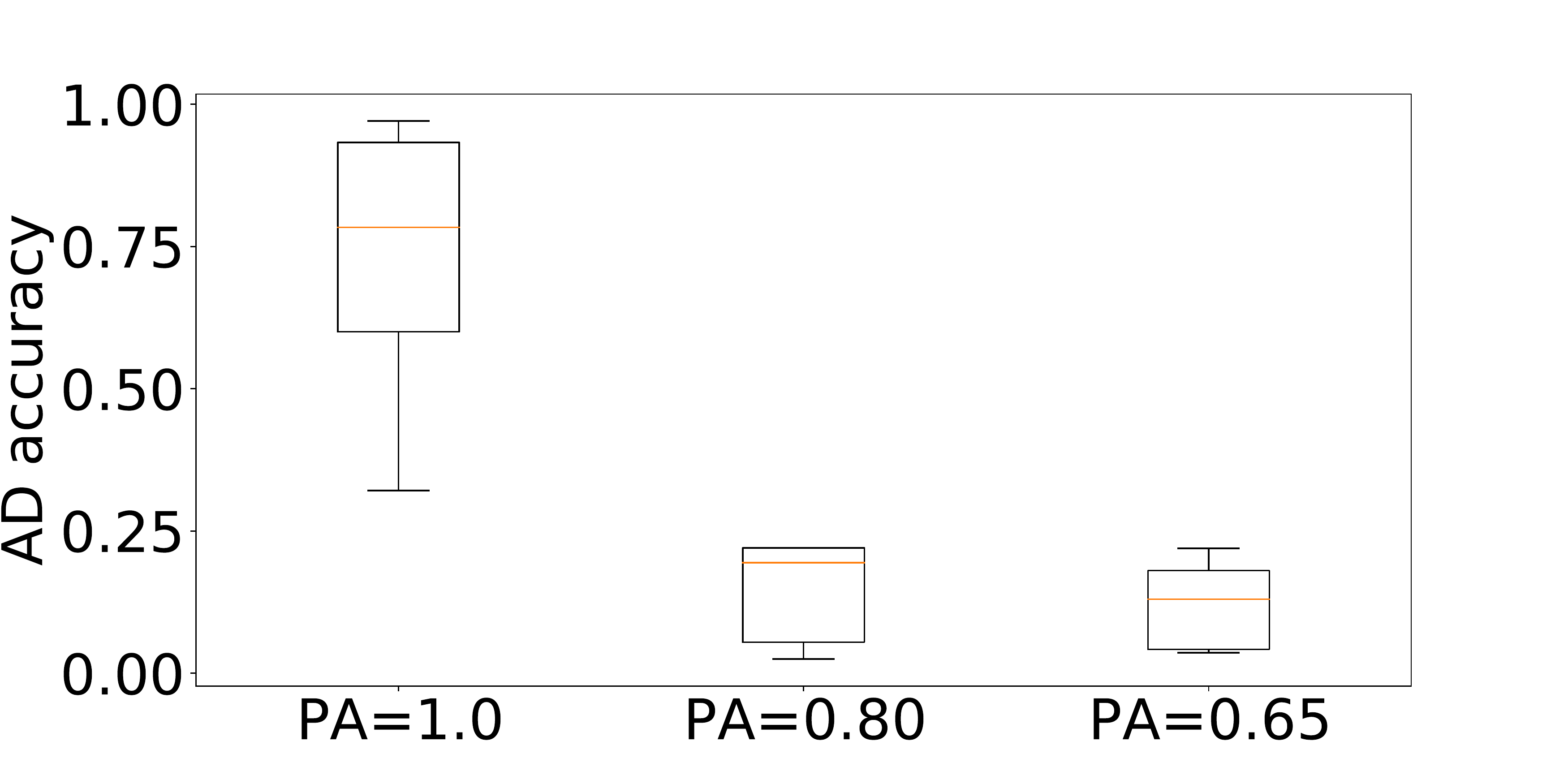}
      \captionof{figure}{Précision de Deeplog et qualité de structuration}
      \label{fig:deeplog}
    \end{minipage}
\end{figure}

Pour les services en ligne de grande envergure, une seule anomalie peut 
impacter des millions d'utilisateurs \citep{liang2021robust}. Pouvoir détecter de tels 
événements en temps réel permet aux équipes de surveillance de limiter leur impact.
Les journaux de \textit{logs} enregistrent une vaste gamme d'événements au sein d'un système informatique
et sont communément utilisés pour détecter les origines d'une panne, analyser l'activité
ou renforcer la sécurité d'une plateforme\citep{moh2016detecting}.
Les \textit{logs} ont prouvé leur valeur pour la détection automatisée d'anomalies
dans plusieurs scénarios \citep{du2017deeplog, zhang2019robust, meng2019loganomaly}.
Afin de ne pas avoir à traiter des données textes brutes, les méthodes
précédemment citées ont recours à une étape de structuration des messages \textit{logs}.
Les \textit{logs} et leur message étant produits par des lignes de code spécifiques (\textit{e.g., print(), logger.log()}),
le processus de structuration vise à retrouver le motif sous-jacent à chaque entrée
(Figure \ref{fig:logparsing}).

La qualité de cette étape de structuration a un impact sur l'efficacité des méthodes en aval.
La Figure \ref{fig:deeplog} illustre l'influence de la qualité de structuration des logs (\textit{Parsing Accuracy (PA)}) sur la précision (\textit{Anomaly Detection (AD) accuracy}) de Deeplog (\cite{du2017deeplog})
une méthode de détection d'anomalies.
Ici, une perte de 0,2 en \textit{PA} réduit de 0,75 la précision \textit{AD} maximale (0,97 à 0,22) et de 0,59 la précision \textit{AD} moyenne (0,78 à 0,19). 
Le graphique présenté est une distribution des résultats de 10 instances de Deeplog entraînées sur différentes 
versions inégalement structurées d'un jeu de données HDFS\footnote{https://github.com/logpai/loghub/tree/master/HDFS}. 
Nous avons utilisé le même ensemble de données et les paramètres décrits dans l'article original présentant Deeplog.

Dans le présent article, nous présentons USTEP, une méthode de structuration en flux des messages \textit{logs} (Section \ref{sec:ustep}). 
Basée sur un arbre de recherche évolutif, USTEP est capable de traiter des messages en temps constant
sans connaissance préalable de l'environnement de \textit{logs}.
La Section \ref{sec:eval} présente une évaluation comparative de notre approche avec quatre autres méthodes de références.
Menée sur 13 ensembles de données issues d'applications différentes,
celle-ci met en valeur la supériorité d'USTEP en matière d'efficacité et de robustesse.
Les travaux présentés ici ont été réalisés en partenariat avec 3DS OUTSCALE, fournisseur de 
services \textit{cloud} qui nous a donné accès à des données internes et de la puissance de calcul.
Nous avons mis en ligne notre implémentation ainsi que 
le code des expérimentations avec leurs paramètres afin de permettre une réutilisation facile de notre travail. 
Les travaux présentés dans ce papier sont un extrait d'une publication 
pour la conférence internationale en fouille de données IEEE ICDM \citep{vervaet2021ustep}.
\section{Structuration des logs en ligne} \label{sec:SOA}
L'intérêt de nombreuses applications pour la structuration des \textit{logs} a
motivé le développement de plusieurs algorithmes \citep{zhu2019tools}.
Notre étude de l'état de l'art a permis d'identifier quatre méthodes de structuration en flux: 
SHISO \citep{mizutani2013incremental} et LenMa \citep{shima2016length} basées sur des approches 
par regroupement de motifs; Spell \citep{du2016spell} basée sur une recherche de
la plus longue séquence commune entre un message et les patrons
déjà découverts; Drain \citep{he2017drain}, une méthode utilisant un arbre de 
partitionnement à profondeur fixe pour encoder des patrons et en découvrir de nouveaux. 
En raison de leur proximité contextuelle avec la nôtre, ce sont les quatre méthodes que
nous avons utilisées dans notre évaluation.

Nous reprenons ici le formalisme introduit par \cite{nedelkoski2020self} et
définissons les \textit{logs} comme des séquences de messages textes 
temporellement ordonnés $\mathcal{L} = (l_i : 1,2,...)$, avec $i$ l'index 
d'un message dans la séquence considérée. Les jetons sont les plus petits
éléments insécables d'un message. Chaque message est constitué d'une séquence finie
de jetons (mots) séparés par des espaces $\mathbf{t_i} = (t_j^i : t\in\mathcal{T},j=1,2,...,\lvert \mathbf{t_i} \rvert)$.
Avec $\mathcal{T}$ l'ensemble des jetons, $j$ la position d'un jeton au sein 
d'un message $l_i$, et $\lvert \mathbf{t_i} \rvert$ le nombre total de jetons dans ce message.
L'éclatement en séquence de jetons est une fonction de transformation $\mathcal{M}: l_i \rightarrow t_i,  \forall i$.
Ici, $\mathcal{M}$ est une séparation sur les espaces de $l_i$.
Le but des opérations de structuration est de séparer les parties constantes des parties
variables d'un message \textit{log}. Nous représenterons ici les patrons et leurs variables
par un couple $EV_i = ((e_i, v_i) : e \in \mathsf{E},i=1,2,...)$.
Avec $\mathsf{E}$ l'ensemble des patrons et $V_i$ la liste des variables de $l_i$. Une méthode de structuration de \textit{log}
est donc assimilable à une fonction $f : l_i \rightarrow EV_i$.

Avec USTEP, nous proposons une méthode robuste et efficace de structuration des \textit{logs} en flux. 
Notre travail est motivé par les besoins
de l'écosystème \textit{cloud}, cependant nous sommes persuadés qu'un grand nombre 
de domaines pourrait en profiter (détection d'anomalies \citep{du2017deeplog},
analyse de la consommation \citep{anitha2016survey}, dissection d’erreurs \citep{lu2017log}).
\section{USTEP: Algorithme} \label{sec:ustep}
Une instance d’USTEP est définie
par $\mathcal{U} = \{\mathcal{P},\sigma,\phi\}$.
Avec, $\mathcal{P} = \{\mathcal{V},\widetilde{\mathsf{E}},\mathcal{E}\}$
un arbre de recherche, $\sigma\in[0;1]$ , et $\phi\in\mathbb{N}^*$ deux paramètres
que nous détaillons ultérieurement.
Ici $\mathcal{V} = \{\upsilon_k\}_{k=1}^\mathsf{N}$
représente l'ensemble des nœuds, $\widetilde{\mathsf{E}}$ l'ensemble des patrons découverts,
et $\mathcal{E}\subset\mathcal{V} \times \widetilde{\mathsf{E}}$
l'ensemble des liaisons nœuds/patrons.

\begin{figure}[h]
    \centering
    \begin{minipage}{.4\textwidth}
      \centering
      \includegraphics[width=0.9\linewidth]{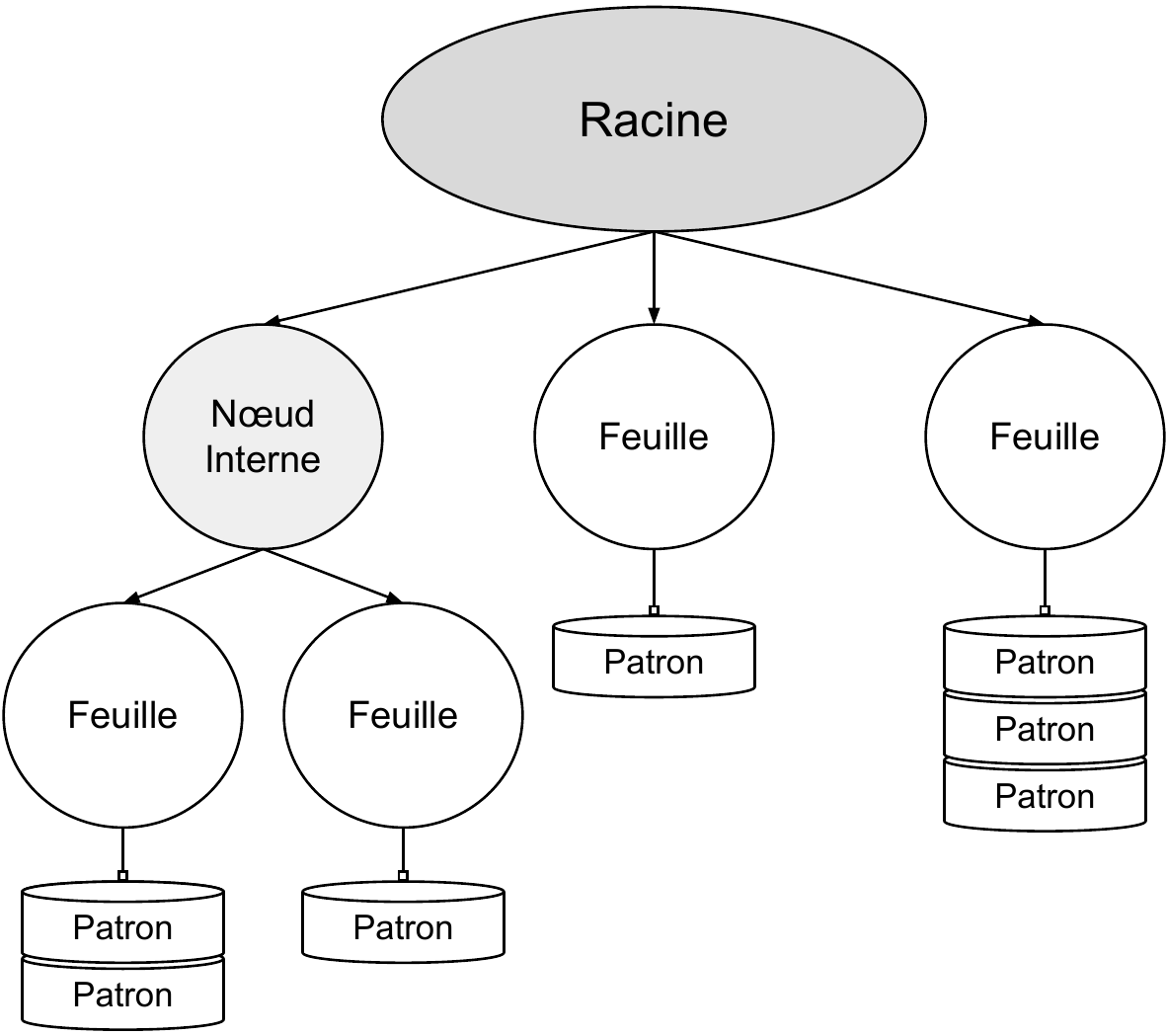}
      \captionof{figure}{Structure de l'arbre}
      \label{fig:tree_structure}
    \end{minipage}%
    \begin{minipage}{.6\textwidth}
      \centering
      \includegraphics[width=0.9\linewidth]{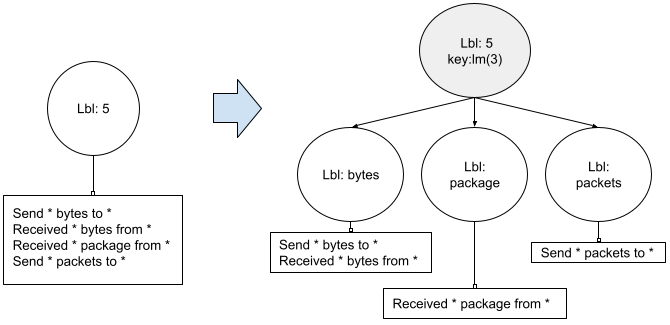}
      \captionof{figure}{Exemple: Eclatement de feuille}
      \label{fig:split_exemple}
    \end{minipage}
\end{figure}

\textbf{Structure de l'arbre:} USTEP utilise un arbre de recherche évolutif 
pour encoder les patrons découverts, les raffiner et en découvrir de nouveaux.
Celui-ci est constitué de quatre types d'éléments (Figure \ref{fig:tree_structure}):
une racine $\upsilon_1$, unique nœud présent à l'initialisation; des nœuds internes
servant à encoder les règles de structuration; des feuilles auxquelles sont attachés
des patrons.

\textbf{Étape préliminaire - Prétraitement:} Les utilisateurs peuvent
fournir des expressions régulières basées sur leur connaissance du contexte pour améliorer le
processus de structuration. Au début de chaque traitement, les jetons concernés 
seront marqués comme emplacement de variable. Cette étape est optionnelle,
mais c'est un moyen d'utiliser sa connaissance du domaine
pour améliorer le processus.

\textbf{Étape 1 - Descente de l'arbre:} 
Les patrons étant attachés aux feuilles, la première étape consiste
à atteindre une feuille. La descente d'arbre est opérée par un mécanisme de 
\textit{key}, \textit{label}. À leur création chaque nœud (interne et feuille)
se voit affecter un label. Plusieurs nœuds peuvent avoir le même label, mais celui-ci 
est unique parmi les enfants d'un même nœud. 
Chaque nœud interne a une méthode $key_{\upsilon_k}(l_i) \rightarrow t_j$, et 
la racine une méthode $key_{\upsilon_1}(l_i) \rightarrow \lvert \mathbf{t_i} \rvert$.
Cette $key_{\upsilon_1}$ étant la première règle de descente, elle
garantit que tous les patrons attachés à une feuille auront le même nombre de jetons.
On commence le processus de descente à la racine ($\upsilon = \upsilon_1$) et on obtient le nœud suivant
grâce à la méthode key du nœud courant $\upsilon$. Cette opération est répétée 
jusqu'à ce que $\upsilon$ soit une feuille. Si aucun nœud fils du nœud 
courant n'a pour label $key_{\upsilon}(l_i)$, on crée une nouvelle feuille avec ce 
label, on l'ajoute aux fils du nœud courant.

\textbf{Étape 2 - Affectation de patron:} À la fin de l'étape précédente, une feuille a été
atteinte. Ici, on recherche un patron représentatif de
$l_i$. Pour ce faire on calcul le facteur de similarité $simF(l_i, \widetilde{e})$
avec tous les patrons $\widetilde{e} \in \widetilde{\mathsf{E}}$ fils de $\upsilon$.
Ce facteur de similarité correspond à la proportion de jetons identiques à même position
entre le message $l_i$ et un patron $\widetilde{e}$. Il est calculé comme suit:

\begin{multicols}{2}
  \begin{equation}
    simF(l_i, \widetilde{e}) = \dfrac{\sum_{j=1}^{\lvert \mathbf{t_i} \rvert} equ(t_j^i, \widetilde{e}(j))}{\lvert \mathbf{t_i} \rvert}
    \label{eq:compute_proximity_factor}
  \end{equation}\break
  \begin{equation}
    equ(t_1, t_2) = 
     \begin{cases}
       1       & \quad \text{if }  t_1 = t_2  \\
       0  & \quad \text{otherwise}
     \end{cases}
   \label{eq:compute_eq}
  \end{equation}
\end{multicols}

Avec $\widetilde{e}(j)$, le $j$-ème jeton de $\widetilde{e}$.
Si la plus grande simF calculée est supérieure à $\sigma$,
on choisit le patron associé pour représenter $l_i$.
Le cas échéant, un nouveau patron égal à $l_i$ est créé et ajouté
à la feuille. 
Si $l_i$ est assimilé à un patron déjà découvert, celui-ci est mis à jour.
Cette mise à jour remplace tous les jetons non identiques à position équivalente 
par des emplacements de variable ('*').

\textbf{Étape 3 - Éclatement des feuilles:}
À la fin de l'Étape 2, un patron $\widetilde{e}$ a été affecté au message log $l_i$.
Afin de perpétuellement augmenter le nombre maximal de calculs de \emph{simF} avec l'augmentation du nombre de patrons découverts:
USTEP transforme les feuilles saturées en nœud interne.
Une feuille est considérée comme saturée si elle est liée à plus de $\phi$ patrons.
Le processus d'éclatement commence par calculer le nombre de jetons différents à chaque position parmi les patrons de $\upsilon$.
La position avec la plus grande diversité est choisie comme pivot $p$.
$\upsilon$ est transformé en nœud interne avec une méthode $key(l_i) = t_p^{i}$.
Pour chaque jeton unique parmi les patrons de $\upsilon$ à la position $p$,
une nouvelle feuille avec ce label est créée et attachée à $\upsilon$.
Les patrons sont alors transférés de $\upsilon$ aux nouvelles feuilles en fonction 
de leurs jetons en position $p$. La Figure \ref{fig:split_exemple} illustre le 
processus de division pour une feuille saturée avec $\phi = 3$.
Dans cet exemple, $p=3$ car c'est la position avec le plus de jetons différents 
parmi les patrons (\textit{bytes}, \textit{packages}, \textit{packets}).

\textbf{Complexité spatiale:} USTEP a une complexité spatiale proportionnelle au nombre de patrons et de nœuds dans $\mathcal{P}$,
soit $\mathcal{O}(\lvert \mathcal{V} \rvert \times \lvert \widetilde{\mathsf{E}} \rvert)$.

\textbf{Complexité temporelle:} La complexité temporelle d’USTEP est de
$\mathcal{O}(\lvert \mathbf{t_i} \rvert - 1 + \lvert \mathbf{t_i} \rvert \times (\phi- 1) + \lvert \mathbf{t_i} \rvert \times (\phi))$,
soit $\mathcal{O}(\lvert \mathbf{t_i} \rvert)$.
Cela représente le cas ou l'on doit faire $\lvert \mathbf{t_i} \rvert - 1$ 
étapes de descente d'arbre, suivies de $(\phi- 1)$ calculs de $simF$.
Le tout menant à une saturation de la feuille trouvée, ce qui demande
$\lvert \mathbf{t_i} \rvert \times \phi$ de plus pour trouver le pivot.
En faisant la supposition raisonnable que $mean(\lvert \mathbf{t_i} \rvert)$ est constant dans le temps,
on a une complexité temporelle en $\mathcal{O}(1)$.
Cette hypothèse est raisonnable puisque la longueur moyenne des \textit{logs} n'est pas vouée
à évoluer avec le fonctionnement du système.
\section{Evaluation}\label{sec:eval}

\begin{table*}[t]
    \centering
    \begin{tabular}{l c c c c c}
      \hline
      Nom & \#messages & taille(B) & \#patrons & \#j/ligne & \%variables\\
      \hline
      Apache & 2 000 & 168K & 6 & 6.3 & 23\%\\
      BGL & 2 000 & 310K & 120 & 6.3 & 25\%\\
      Hadoop & 2 000 & 376K & 114 & 8.4 & 37\%\\
      HDFS & 2 000 & 282K & 14 & 7.4 & 45\%\\
      HPC & 2 000 & 148K & 46 & 3.5 & 18\%\\
      Mac & 2 000 & 312K & 341 & 9.4 & 28\%\\
      OpenSSH & 2 000 & 220K & 27 & 8.7 & 28\%\\
      OpenStack & 2 000 & 582K & 43 & 9.0 & 45\%\\
      Thunderbird & 2 000 &  318K & 149 & 8.5 & 16\%\\
      Zookeeper & 2 000 & 274K & 50 & 6.3 & 18\%\\
      HDFS-2 & 11 175 629 & 1.5G & 30 & 7.4 & N.A\\
      OpenStack-2 & 207 820 & 54M & 43 & 9.0 & N.A\\
      Internal-1 & 1 750 916 & 1.2G & N.A & 28.2 & N.A\\\hline
   \end{tabular}
   \caption{Caractéristiques des jeux de données}
   \label{tab:datasets}
   \end{table*}

  \textbf{Jeux de données:} Afin d'évaluer notre approche, nous avons sélectionné 13 jeux de données
  provenant d'applications différentes. On retrouvera parmi eux des \textit{logs} venant
  de systèmes distribués (\textit{HDFS, OpenStack, Zookeeper, Internal-1}),
  d'applications serveur (\textit{OpenSSH}), de supercalculateurs (\textit{BGL, HPC, Thunderbird})
  et d'opérateurs système (\textit{Mac}). À l'exception d'Internal-1 qui est issu 
  de systèmes internes 3DS OUTSCALE, tous ces jeux sont disponibles en ligne.
  Leur origine et leur méthode de collection sont détaillées dans les travaux de \cite{he2020loghub}.
  Le Tableau \ref{tab:datasets} regroupe les caractéristiques de ces différents jeux de données avec:
  le nombre de lignes de \textit{logs} (\emph{\#messages}); la taille mémoire; le nombre réel de patrons
  sous-jacents (\emph{\#patrons}); le nombre moyen de jetons par ligne (\emph{j/ligne}) et 
  la proportion de jetons variables dans le jeu (\emph{\%variables}).

  \textbf{Contexte expérimental:}  
  Nous comparons ici les performances d’USTEP avec celles de quatre autres méthodes
  de structuration des logs en lignes identifiées dans notre état de l'art (Drain, LenMa, SHISO, Spell).
  Par mesure d'équité, les mêmes expressions régulières ont été appliquées pour le 
  prétraitement de chaque méthode. 
  Les expériences ont été réalisées sur une machine virtuelle \textit{cloud} de type
  CentOS Linux 7.8 avec 32 cœurs et 62 GB de RAM.
  Par manque de place, nous ne détaillons pas ici les valeurs utilisées
  pour les hyperparamètres de chaque méthode, mais ceux-ci sont disponibles en ligne avec
  le code source des expériences.

  \begin{table}[ht]
    \centering
  \begin{tabular}{l c c c c c}
    \hline
     Jeu de donnée & Drain & LenMa & SHISO & Spell & USTEP\\
     \hline
     Apache & \textbf{1} &  \textbf{1} &  \textbf{1} &\textbf{1} &  \textbf{1} \\
     BGL & 0.963 & 0.690 & 0.711 & 0.787 & \textbf{0.964}\\
     Hadoop & 0.948  & 0.885 & 0.867 & 0.778 & \textbf{0.951}\\
     HDFS & 0.998 & 0.998 & 0.998 & \textbf{1} & 0.998\\
     HPC & 0.887 & 0.830 & 0.325 & 0.654  & \textbf{0.906}\\
     Mac & 0.787 & 0.698 & 0.595 & 0.757 & \textbf{0.848}\\
     OpenSSH & 0.788 & 0.925 & 0.619 & 0.554 & \textbf{0.996}\\
     OpenStack & 0.733 & 0.743 & 0.722 & \textbf{0.764} & \textbf{0.764}\\
     Thunderbird & \textbf{0.955} &0.943 & 0.576 & 0.844 & 0.954\\
     Z.keeper & 0.967 & 0.841 & 0.660 & 0.964 & \textbf{0.988}\\\hline
     Moyenne & 0.903 & 0.855& 0.707 & 0.810 & \textbf{0.937}\\\hline
  \end{tabular}
  \caption{Précision des méthodes de structuration}
  \label{tab:PA_evaluation}
  \end{table}
  
  \textbf{Efficacité de la structuration:}
  Les méthodes de structuration des \textit{logs} sont usuellement évaluées
  grâce à la précision de structuration (PA) (\cite{zhu2019tools,du2016spell}).
  Celle-ci représente le ratio de messages correctement structurés par rapport au nombre total de 
  messages. Un message est considéré comme correctement structuré s’il est associé au même
  groupe que son patron sous-jacent. 

  Pour évaluer l'efficacité des méthodes sélectionnées, nous avons utilisé nos 10 
  jeux labélisés. Afin d'éviter de possible biais, les hyperparamètres de chaque méthode ont été sélectionnés en répétant 
  100 fois l'expérience avec des valeurs différentes et en conservant le meilleur résultat (Tableau \ref{tab:PA_evaluation}).
  Notre méthode USTEP obtient la meilleure PA moyenne à 0.937 PA et la plus haute PA pour 8 des 10 jeux considérés tout
  en étant très proche du meilleur résultat pour les deux derniers jeux de données.

\begin{figure}[t]
    \centering
    \begin{minipage}{.5\textwidth}
      \centering
      \includegraphics[width=0.9\linewidth]{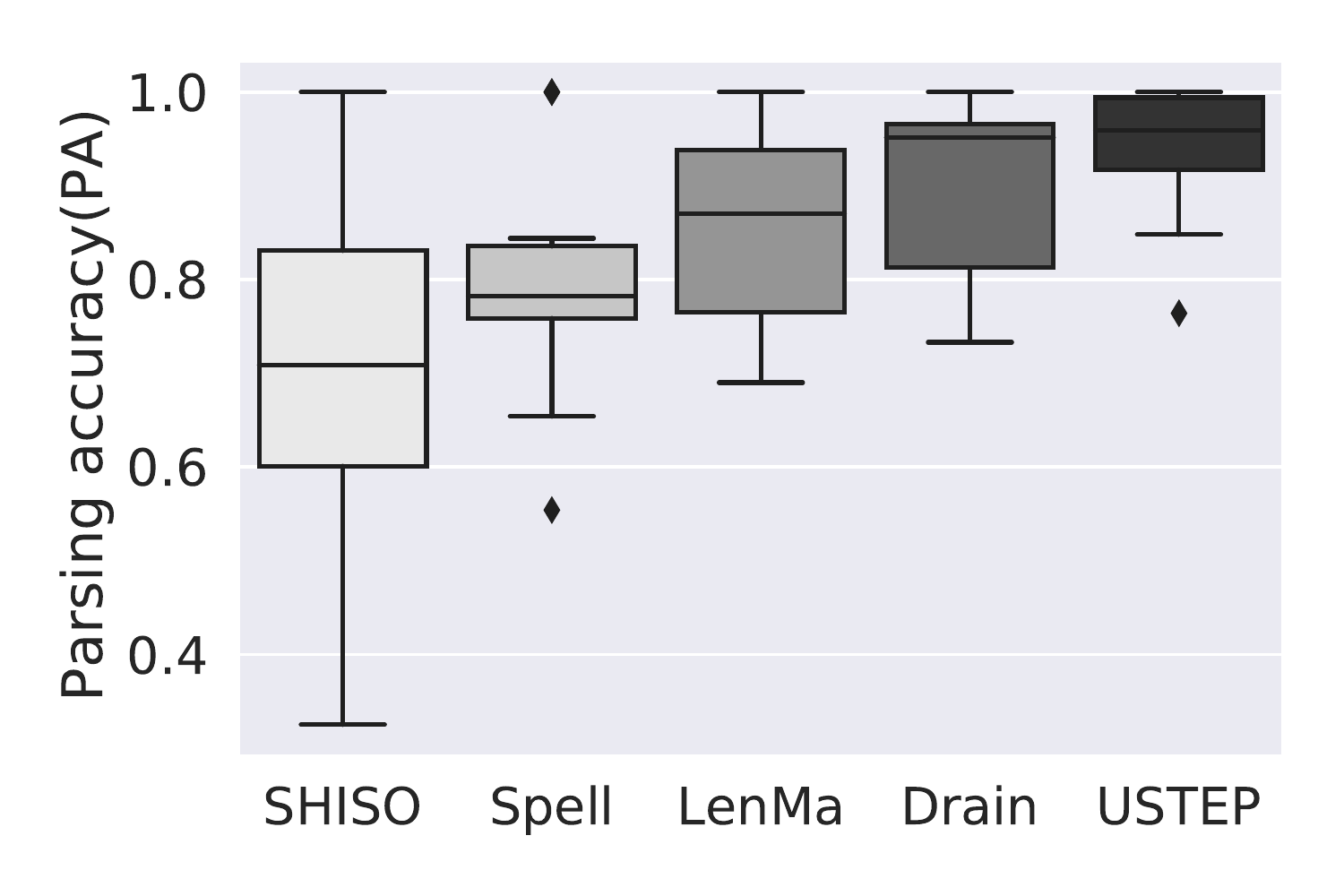}
      \captionof{figure}{Robustesse de structuration}
      \label{fig:robustness}
    \end{minipage}%
    \begin{minipage}{.5\textwidth}
      \centering
      \begin{subfigure}{.45\linewidth}
        \centering
        \includegraphics[width=\linewidth]{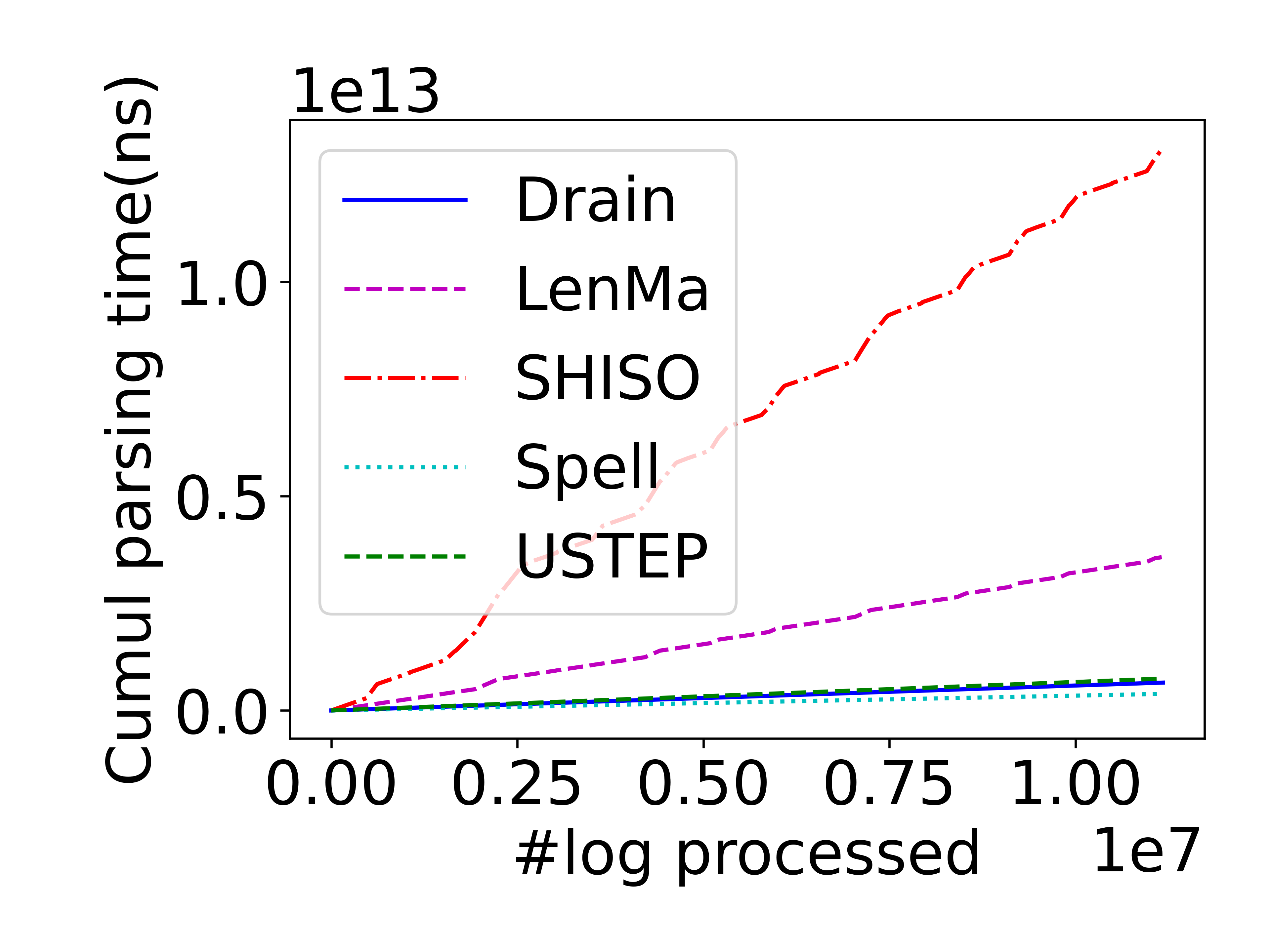}
        \caption{HDFS-2}
      \end{subfigure}
      \begin{subfigure}{.45\linewidth}
        \centering
        \includegraphics[width=\linewidth]{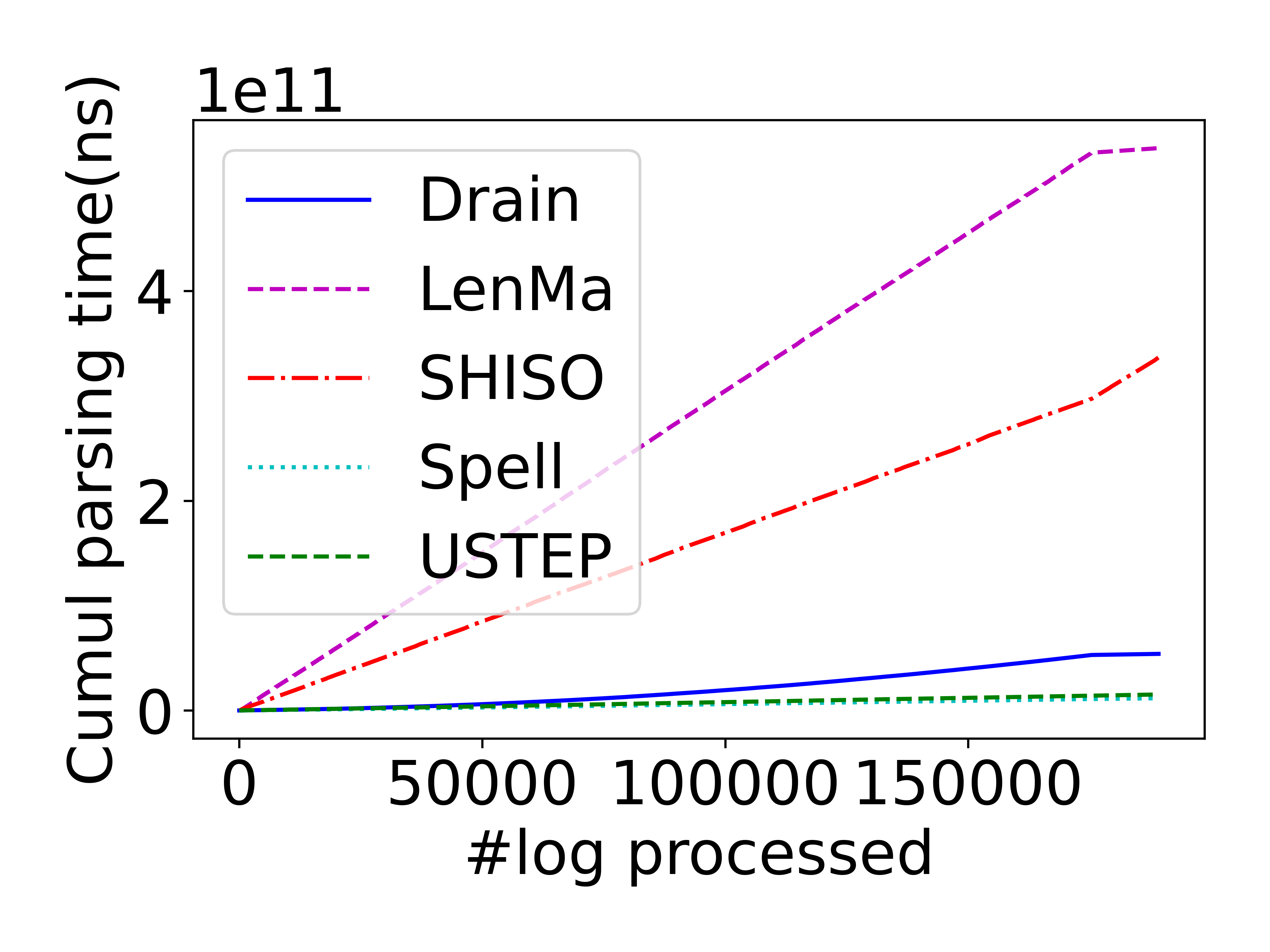}
        \caption{OpenStack-2}
      \end{subfigure}
      \begin{subfigure}{.45\linewidth}
        \centering
        \includegraphics[width=\linewidth]{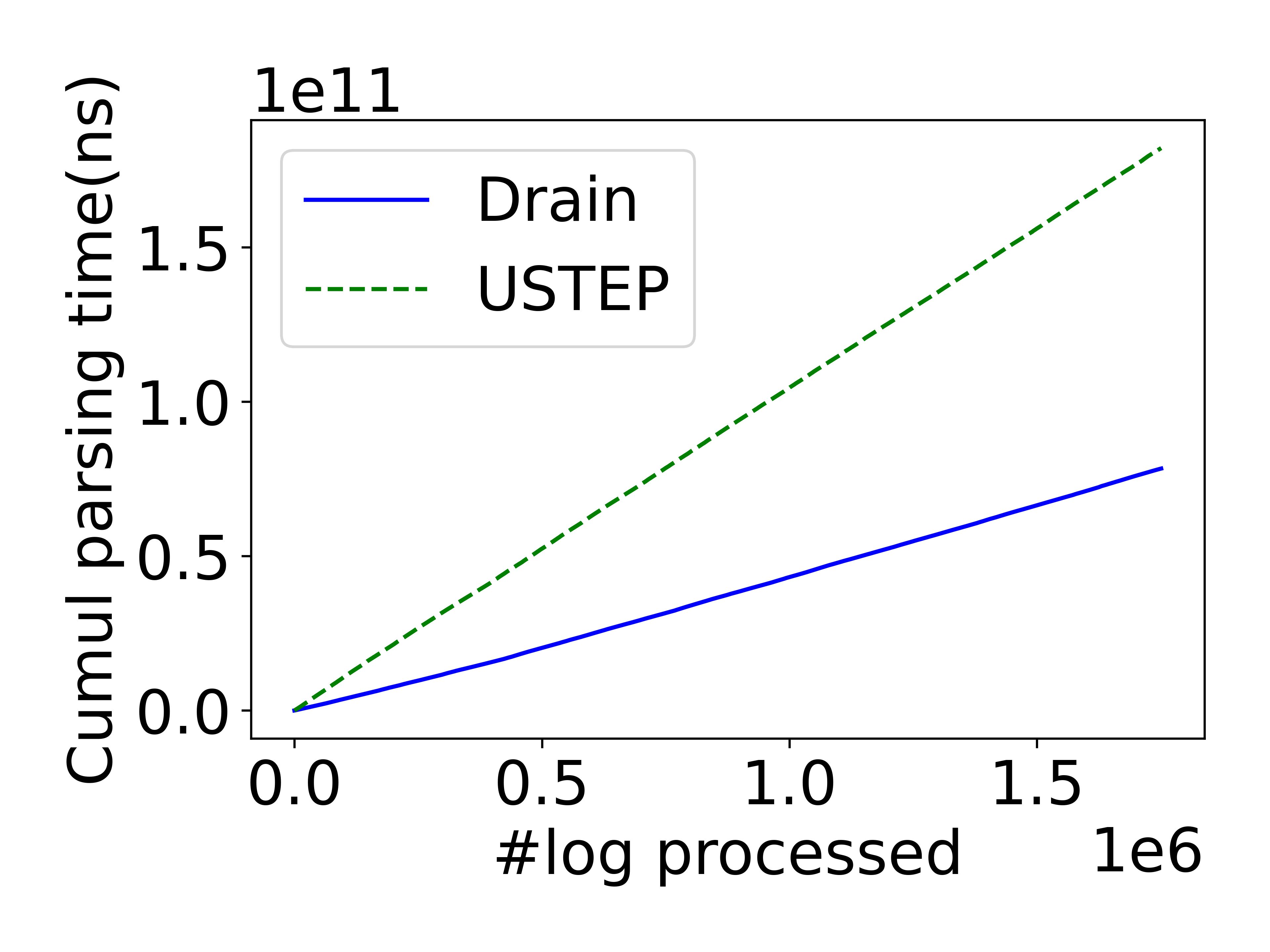}
        \caption{Internal-1}
      \end{subfigure}
      \captionof{figure}{Vitesse de structuration}
      \label{fig:parsing_speed}
    \end{minipage}
\end{figure}

  \textbf{Robustesse de la structuration:} Nous sommes persuadés que pour 
  être utilisée en production, une méthode de structuration de \textit{logs}
  doit être robuste et maintenir ses performances sur des jeux de données provenant
  de systèmes divers. La Figure \ref{fig:robustness} représente une distribution 
  en boite de la PA obtenue par chaque méthode sur les différents jeux de données.
  Les méthodes sont ordonnées par médiane croissante. SHISO termine en bas de l'échelle
  et USTEP en haut. La distance du premier au troisième quartile est de 0.23 pour SHISO,
  0.146 pour Drain, 0.079 pour Spell et 0.073 pour USTEP.
  Couplée au fait qu'il ait la meilleure PA moyenne (0.959), la faible variance d’USTEP
  la rend plus généralisable que les autres méthodes considérées.
  Cette capacité de généralisation est particulièrement intéressante pour notre contexte \textit{cloud}
  dans lequel la base de code est soumise à de nombreux changements et où le client 
  est maître des applications qu'il souhaite faire tourner et de leurs versions.

  \textbf{Vitesse de structuration:}
  Structurer les \textit{logs} n'est généralement qu'une étape préliminaire pour 
  des tâches d'inspection et peut donc devenir un goulot d'étranglement
  temporel. Pour mesurer la capacité de chaque méthode à traiter rapidement des grands volumes de 
  \textit{logs}, nous avons utilisé nos trois jeux de données (\textit{HDFS-2}, \textit{OpenStack-2}, \textit{Internal-1})
  et enregistré le temps mis par chaque méthode pour les traiter complètement.

  Les trois graphiques de la Figure \ref{fig:parsing_speed} affichent l'évolution
  du temps total de traitement relativement au nombre de \textit{logs} traités.
  Drain et USTEP se présentent comme les méthodes les plus rapides avec un temps de
  structuration constant.
  Seules ces deux méthodes ont réussi à traiter \textit{Internal-1} dans un temps
  avoisinant les 5 heures. Pour ce jeu de données, Spell, SHISO et LenMa n'avaient pas obtenu de résultat
  en moins d'une semaine. Spell présente une vitesse constante sur \textit{HDFS-2} et 
  \textit{OpenStack} et n'arrive pas à structurer le dernier jeu de données en un temps raisonnable.
  Plusieurs facteurs expliquent ce résultat: 1) Spell se base sur l'algorithme de recherche 
  de la plus longue séquence commune avec une complexité en $\mathcal{O}(\lvert \mathbf{t_i} \rvert \times m)$,
  or \textit{Internal-1}   a un plus grand nombre moyen de jetons par ligne (28.2) $\lvert \mathbf{t_i} \rvert$
  que les autres jeux (7.4 et 9.0); 2) \textit{Internal-1} est le seul jeu pour lequel aucune
  expression régulière n'a été utilisée, ce qui complique la tâche de Spell qui se base sur de nombreuses
  optimisations pour améliorer sa complexité moyenne.

\section{Conclusion}\label{sec:conclusion}
La structuration des messages \textit{logs} est une étape obligatoire pour de nombreuses
méthodes d'analyse des logs. Dans cet article, nous avons présenté USTEP,
une méthode de structuration en ligne basée sur un arbre de recherche évolutif. Notre évaluation
sur 13 jeux provenant d'applications différentes met en valeur le temps
de traitement constant ainsi que la supériorité en termes de précision (+3.4\%  à93.7\%) et de robustesse de notre approche
par rapport  à quatre autres méthodes de l'état de l'art. Dans le futur, nous voulons améliorer la mise à l’échelle
de notre méthode au travers d'une étude sur sa distribution sur plusieurs machines. 
Nous souhaitons également utiliser la sortie d’ USTEP pour faire progresser la recherche en détection automatisée
des anomalies appliquée aux \textit{logs}.
\\

\noindent\textbf{Remerciements:} Nous remercions 3DS OUTSCALE et l'ANRT qui contribuent au financement de nos travaux de recherches 
au travers d'un programme CIFRE (Convention 2020/0289).

\bibliographystyle{rnti}
\bibliography{biblio}

\appendix

\Fr

\end{document}